\documentclass[letterpaper]{article} % DO NOT CHANGE THIS
\usepackage{aaai24}  % DO NOT CHANGE THIS
\usepackage{times}  % DO NOT CHANGE THIS
\usepackage{helvet}  % DO NOT CHANGE THIS
\usepackage{courier}  % DO NOT CHANGE THIS
\usepackage[hyphens]{url}  % DO NOT CHANGE THIS
\usepackage{graphicx} % DO NOT CHANGE THIS
\usepackage{natbib}  % DO NOT CHANGE THIS AND DO NOT ADD ANY OPTIONS TO IT
\usepackage{caption} % DO NOT CHANGE THIS AND DO NOT ADD ANY OPTIONS TO IT
\usepackage{amsfonts}
\usepackage{amsthm}
\usepackage{amsmath}
\usepackage{amssymb}
\usepackage{newtxmath}
\usepackage{multirow}
\usepackage{color}
\usepackage{float}

\DeclareMathOperator*{\argmin}{arg\,min}
\usepackage{newfloat}
\usepackage{listings}
\DeclareCaptionStyle{ruled}{labelfont=normalfont,labelsep=colon,strut=off} % DO NOT CHANGE THIS
\floatstyle{ruled}
\newfloat{listing}{tb}{lst}{}
\floatname{listing}{Listing}
\title{ChartDETR: A Multi-shape Detection Network for Visual Chart Recognition}
\author{
    Wenyuan Xue\textsuperscript{\rm 1}, 
    Dapeng Chen\textsuperscript{\rm 1},
    Baosheng Yu\textsuperscript{\rm 2},
    Yifei Chen\textsuperscript{\rm 1},
    Sai Zhou\textsuperscript{\rm 1},
    Wei Peng\textsuperscript{\rm 1}
}
\affiliations{
    \textsuperscript{\rm 1}IT Innovation and Research Center, Huawei Technologies Ltd.\\
    \textsuperscript{\rm 2}The University of Sydney\\
    \{xuewenyuan1,chendapeng8\}@huawei.com, baosheng.yu@sydney.edu.au
}

\usepackage{bibentry}
\begin{document}

\maketitle

\begin{abstract}
Visual chart recognition systems are gaining increasing attention due to the growing demand for automatically identifying table headers and values from chart images. Current methods rely on keypoint detection to estimate data element shapes in charts but suffer from grouping errors in post-processing. To address this issue, we propose ChartDETR, a transformer-based multi-shape detector that localizes keypoints at the corners of regular shapes to reconstruct multiple data elements in a single chart image. Our method predicts all data element shapes at once by introducing query groups in set prediction, eliminating the need for further postprocessing. This property allows ChartDETR to serve as a unified framework capable of representing various chart types without altering the network architecture, effectively detecting data elements of diverse shapes. We evaluated ChartDETR on three datasets, achieving competitive results across all chart types without any additional enhancements. For example, ChartDETR achieved an F1 score of 0.98 on Adobe Synthetic, significantly outperforming the previous best model with a 0.71 F1 score. Additionally, we obtained a new state-of-the-art result of 0.97 on ExcelChart400k. The code will be made publicly available.
% Visual chart recognition systems have gained significant interest in recent years due to the growing need to identify table headers and their values from chart images automatically. Current deep learning methods typically rely on keypoint detection to detect the complex shapes of data elements in charts, which can lead to grouping errors when attempting to identify and categorize data elements. To address this issue, we introduce ChartDETR, a transformer-based multi-shape detector that localizes keypoints at the corners of regular shapes to reconstruct multiple data elements in one chart image. Our novel approach doesn't rely on diverse and complex post-processing grouping and instead predicts all the shapes of data elements at once by introducing query groups in set prediction. ChartDETR offers a unified framework that can represent different types of charts without changing the network architecture and effectively detect data elements of various shapes in charts. We evaluated ChartDETR on three datasets and achieved competitive results across all chart types without relying on any bells or whistles. For example, ChartDETR achieved an F1 score of 0.98 on Adobe Synthetic– significantly better than the previous best model with a 0.71 F1 score. We also obtained a new state-of-the-art result of 0.96 on ExcelChart400k. Code will be public.
\end{abstract}

\section{Introduction}

\begin{figure}[t]
    \centering
    \includegraphics[width=\linewidth]{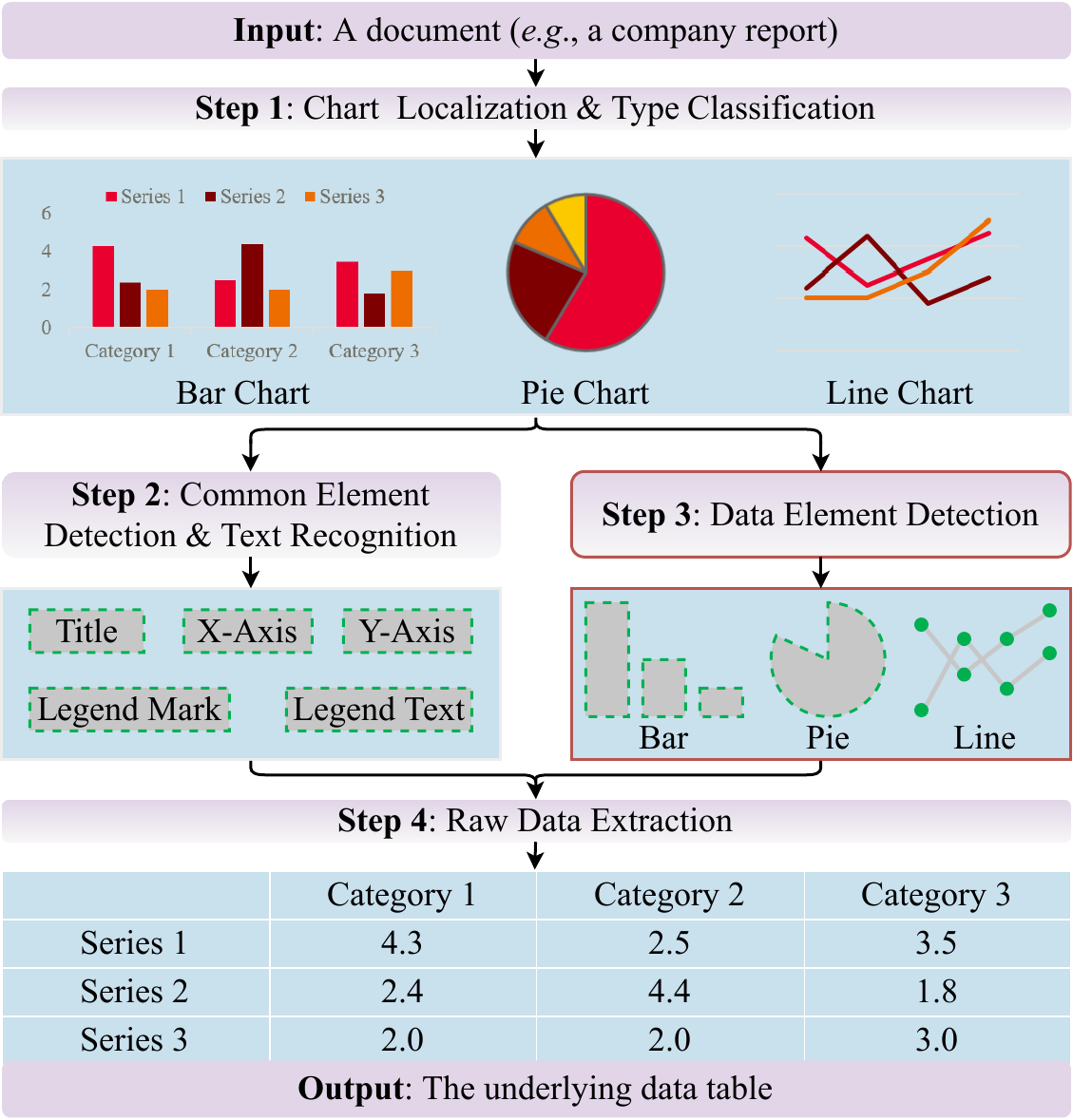}
    \caption{A chart recognition example shows the main concern of this paper. Before extracting raw data from the detected chart area on a document image, the chart recognition system detects various visual elements. Though the general bounding box detection can be directly used for common elements (\emph{e.g.}, the title,  x/y axis, and legend), it is challenging to detect data elements (\emph{e.g.}, bar, line, and pie) due to the different shape types and diverse visual appearances.}
    \label{fig:cmp}
    \vspace{-3mm}
\end{figure}

% Charts are an effective way to analyze important data and make informed decisions. They can be found in various real-world scenarios, including scientific papers, company reports, and health dashboards~\cite{covid19who, huang2007system, liu2013review}. In recent years, there has been growing interest in visual chart recognition systems that can automatically identify table headers and their values from chart images. Given the large number of charts published as images and the fact that raw data is often inaccessible, a visual chart recognition system has the potential to unlock new functions based on chart images. These functions include chart question answering, chart summarization~\cite{masry2022chartqa,kantharaj2022chart}, infographics retrieval~\cite{li2022structure}, and chart redesign~\cite{li2022structure}.

The continuous development of data perception and the widespread use of the internet have led to a vast amount of data being generated and collected. Charts serve as a universal tool for visualizing this data, offering a concise and intuitive representation of complex datasets that allows users to easily identify trends and patterns. However, chart data is often stored in the form of images. To enable machines to comprehend chart images, it becomes essential to convert them into corresponding table data. This conversion proves to be especially beneficial when combined with large language models (LLM)~\cite{masry2022chartqa,kantharaj2022chart,liu-etal-2023-deplot}. By doing so, we enhance our understanding of the chart's underlying meaning and ultimately improve work efficiency.

% %from background to problem

A visual chart recognition system typically consists of multiple steps, as shown in Figure~\ref{fig:cmp}. When a document image containing a chart is provided, the system begins by localizing its position and determining its type. It then proceeds with two parallel steps: one to recognize the common chart elements (such as the title, x/y axis, and legend), and another to detect the data elements. Finally, the system combines the common and data elements to retrieve the table data corresponding to the chart.

Detecting data elements is a crucial but less exploited step in visual chart recognition. However, it remains challenging due to several factors. First, different types of chart elements have significantly different shapes. Second, even chart elements of the same type can possess diverse visual appearances. To address these challenges, current methods detect all keypoints in the given chart image. They either use heatmaps to represent different keypoints or utilize visual transformers~\cite{shivasankaran2023lineex} to directly predict a fixed number of keypoints. One limitation of keypoint detection is its inability to determine whether different points belong to the same data element. Therefore, a hand-designed grouping process is necessary to fully identify and categorize data elements in any given chart. Since different shapes of charts require different grouping rules, this process inevitably introduces grouping errors.

In this paper, we present ChartDETR, a transformer-based multi-shape detector that localizes the keypoints of multiple data elements in one chart image. It inherits from the DETR that gets rid of the post-processing procedure. Compared with DETR, ChartDETR can better handle more complex shapes by using multiple queries rather than one query for each data element. This is because one query with the multi-head attention focuses on features in a local area~\cite{carion2020end, zhu2020deformable}, which is not suitable for charts that may span a wide range of an image. 

Instead of explicitly figuring out which keypoints belong to the same data element, we introduce groups for queries. Each group of queries will induce a group of keypoints. Suppose there are at most $M$ shapes in an image, with each shape requiring $N$ keypoints to describe it. ChartDETR predicts a set of $M \times N$ keypoints. During training, we match each group of $N$ keypoints to one of the ground-truth shapes. Note that not all shapes can be described by a fixed number of keypoints, so $N$ should be larger than the number of ground truth keypoints. To enable point-to-point supervision over the predicted keypoints, a simple duplicated keypoint matching is introduced, which interpolates nearest neighbors to ground truth points during training. The sum of the point-wise loss is also used in a bipartite matching between the ground truth shapes and the predicted keypoint groups.

ChartDETR is a simple yet highly effective method for detecting data elements of charts. The proposed approach has the potential to be generalized to detect any regular/irregular object whose shape can be represented by keypoints. We evaluated our approach on three datasets and achieved competitive results across all chart types. For example, on Adobe Synthetic~\cite{davila2019icdar}, ChartDETR achieved an F1 score of 0.98 – significantly better than the previous best model with a 0.71 F1 score. On ExcelChart400k~\cite{luo2021chartocr}, we obtained a new state-of-the-art result of 0.97. ChartDETR makes three main contributions to the field of chart recognition:
\begin{itemize}
\item Utilizing keypoints, ChartDETR is a unified framework that can represent different types of charts. 
% without changing the network architecture.
\item Introducing query groups and set prediction, ChartDETR predicts all data elements in one shot, eliminating the need for post-processing.
\item Effective and straightforward, ChartDETR achieves state-of-the-art results for bar, line, and pie charts without whistles and bells.
\end{itemize}

\section{Related Work}

\subsection{Data Element Detection in Chart}

\begin{figure}
    \centering
    \includegraphics[width=\linewidth]{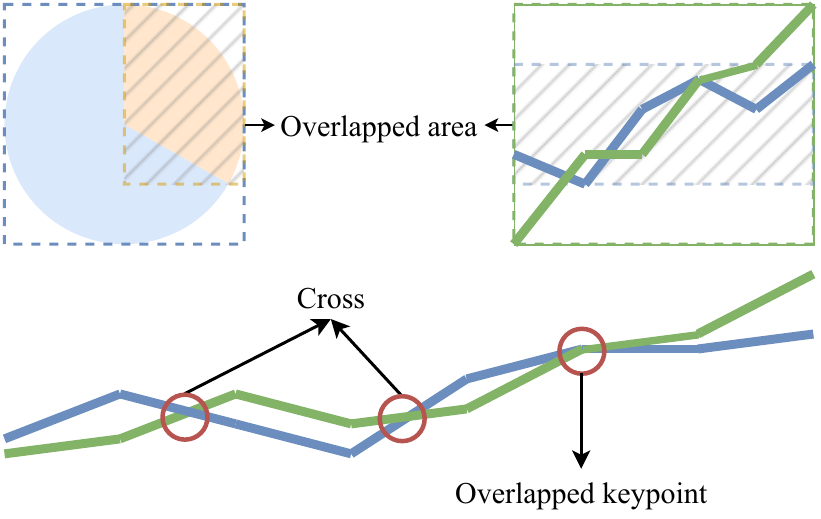}
    \caption{Challenges for pie and line detection. It is hard to represent pies and lines with bounding boxes because large overlapped areas exist. Moreover, different lines may intersect at a crossing and share the same keypoint.}
    \label{fig:challenge}
\end{figure}

Chart images typically comprise lines and shapes, which appear very different from natural images. Early methods ~\cite{balaji2018chart,gao2012view,poco2017reverse,savva2011revision} utilized defined rules to extract the shapes by finding the raw components by color continuity and edge features. However, hand-crafted rules can only work efficiently in constrained settings. For example, ReVision~\cite{savva2011revision} may fail to detect small or rarely-occurring bars. Recent works have attempted to detect chart elements using deep neural networks. For bar charts, a natural approach is to use general object detection to detect bounding boxes of bars~\cite{choi2019visualizing,liu2019data}. As shown in Figure~\ref{fig:challenge}, for pie sectors and lines with knee-points, the shapes cannot be described by bounding boxes but can be represented by keypoints, which help to recover the complete shape. To detect these keypoints for a chart, ChartOCR~\cite{luo2021chartocr} and CHARTER~\cite{shtok2021charter} use heatmap-based approaches to detect different keypoints of various shapes. 
% such as top-left/bottom-right corners of bars, knee-points of lines, and centers/arc-intersections of pies.
LINEEX~\cite{shivasankaran2023lineex} further utilizes a transformer architecture to detect the keypoints. Most of these methods use the bottom-up framework and need sophisticated post-processing methods to group them into different shapes.
% Since a chart image may contain multiple data elements, the detected keypoints do not contain information about which data element they belong to. Therefore, these methods still need sophisticated post-processing methods to group them into different shapes.

\subsection{Transformer based Detector}
Transformer is successfully applied for object detection in natural images. DETR~\cite{carion2020end} is the first transformed-based end-to-end object detector. It is known for not requiring many hand-designed components like anchor design and non-maximum suppression. Many advanced extensions have been developed based on DETR. Deformable-DETR~\cite{zhu2020deformable} proposed the multi-scale deformable self/cross-attention module to improve the training efficiency.
% that only attends a small set of key sampling points around a reference point and achieves better performance than DETR. 
% Efficient DETR~\cite{yao2021efficient} selects top K positions from the encoder's dense prediction to enhance decoder queries. DAB-DETR~\cite{liu2022dab} further extends 2D anchor points to 4D anchor box coordinates to represent queries and dynamically update boxes in each decoder layer. 
DN-DETR~\cite{li2022dn} speeds up DETR training by introducing an auxiliary denoising training task. Then, DINO~\cite{zhang2022dino} improved DN-DETR by using a contrastive way for denoising training, a mixed query selection method for anchor initialization, and a look forward twice scheme for box prediction. For the low training efficiency with few positive queries, H-DETR~\cite{jia2022detrs} adopts a hybrid matching scheme that combines the original one-to-one matching branch with another one-to-many matching during training. Inspired by these DETR-style detectors, we propose ChartDETR for chart images. However, the data elements can have very long and thin shapes, \emph{e.g.}, long bars, and wide lines. A single query cannot account for the whole data element. Therefore, we introduce "groups" for queries so that groups of queries together induce the keypoints to represent the data elements. 

\subsection{Keypoint Detection for Multiple Instances}
Keypoint detection is widely studied for pose estimation, which requires locating multiple critical points of the presented persons from an image. The related techniques can be divided into three categories: the top-down approaches, the bottom-up methods, and the single-stage methods. For multiple instances keypoint detection, the top-down methods~\cite{newell2016stacked,sun2019deep} first employ an object detector to obtain the bounding box of each person instance and then estimate the single-person pose from the cropped area. The bottom-up methods~\cite{kreiss2019pifpaf,papandreou2018personlab} detect all the keypoints in an instance-agnostic fashion, then group them into individuals. In the chart image,  bounding boxes of different data elements may overlap as shown in Figure~\ref{fig:challenge}, which is unsuitable for top-down methods. Therefore, many previous chart recognition methods~\cite{luo2021chartocr,shtok2021charter} utilize the bottom-up pipeline for shape keypoints detection.
% and employ post-processing for data element grouping. 
The two-stage strategy inevitably introduces intermediate errors. PETR~\cite{shi2022end} is a single-stage multi-person pose estimation framework that also adopts a DETR-style framework. For an input image, PETR designs the hierarchical decoders to predict $N$ poses with $K$ joints for each, which significantly increase the computational complexity.
% first predicts $N$ poses with $K$ joints for each and then refines the joints by a joint decoder. The hierarchical decoders significantly increase the computational complexity and the initial joints may not be precise because they are all based on one pose query. 
While our ChartDETR has a more clean architecture that predicts all shape keypoints in one shot and groups keypoints by pivot queries.

\section{Methodology}

\subsection{Shape Representation}
\label{sec:shape_rep}

ChartDETR utilizes keypoints to represent the shape of data elements. This paper considers pie, bar, and line charts.\\

\noindent \textbf{Pie Chart.} A pie chart can be composed of multiple pie sectors. We use the three keypoints for each pie sector to represent its shape: one center point of the circle represents the sector, and the other two endpoints of the arc define the sector's boundaries. \\
% Using these three keypoints for each sector, we can accurately calculate the proportionality of each category in the pie chart.\\

\noindent \textbf{Bar Chart.} In a bar chart, each rectangle represents a particular value or set of values. These rectangles can be arranged horizontally or vertically, depending on the type of chart being used. To represent the shape of each rectangle, we utilize two keypoints typically located at the top-left and bottom-right corners of the rectangle. \\
% By analyzing the rectangles in one chart, we can determine the directions of the bars and extract the key values from the bar chart according to the shape and numbers on the axis.\\

\begin{figure}
    \centering
    \includegraphics[width=\linewidth]{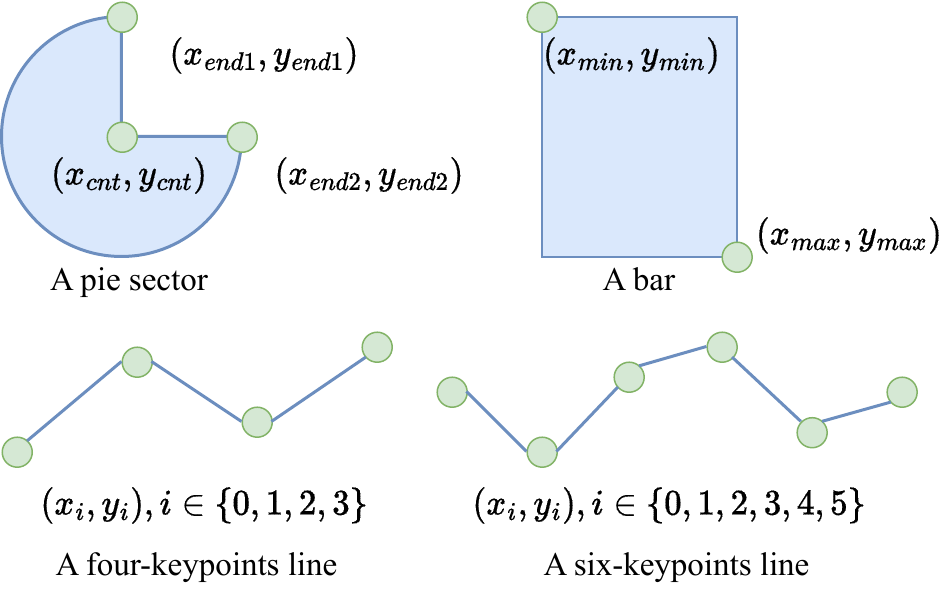}
    \caption{Examples of shape representation for a pie sector, a bar, and two lines with four and six keypoints, respectively.}
    \label{fig:shape_rep}
    \vspace{-3mm}
\end{figure}

\noindent \textbf{Line Chart.} A line chart connects discrete data points with line segments to compose a continuous line. Knee points on this line usually correspond with the source data points. Therefore, we use a set of knee points to represent its shape. As shown in Figure~\ref{fig:shape_rep}, representing a data element in a line chart differs from pies and bars: the number of keypoints for lines varies in different charts. It significantly increases the difficulty of the detection, which usually has a fixed number of outputs. Our ChartDETR will provide a novel way to detect a variable number of keypoints for each instance. 

% \begin{figure}
%     \centering
%     \includegraphics[width=\linewidth]{figs/bar-pie_rep.pdf}
%     \caption{Examples of shape representation for a pie sector and a bar. A pie sector is represented by the center point and two boundary points, while a bar is represented by the top-left point and right-bottom point.}
%     \label{fig:bar-pie_rep}
% \end{figure}

% \begin{figure}
%     \centering
%     \includegraphics[width=\linewidth]{figs/line_rep.pdf}
%     \caption{Examples of two lines with four and six keypoints, respectively.}
%     \label{fig:line_rep}
% \end{figure}

% \subsection{Network Architecture}
\subsection{ChartDETR}
\label{sec:method}

\begin{figure*}
    \centering
    \includegraphics[width=\linewidth]{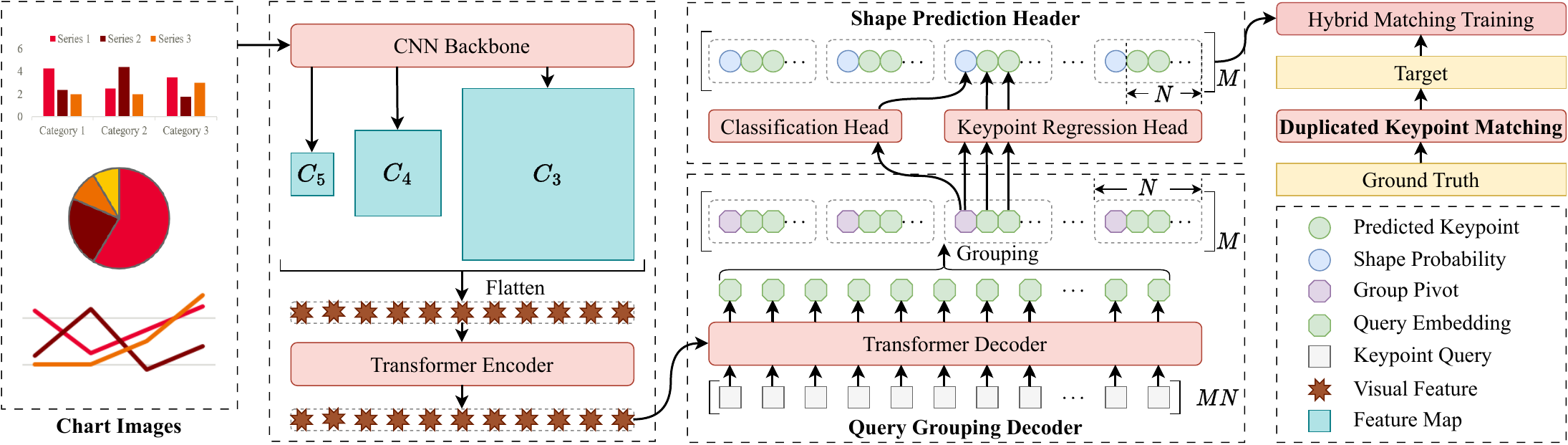}
    \caption{The main framework of the proposed multi-shape detection network or ChartDETR.}
    \label{fig:chart_detr}
    \vspace{-3mm}
\end{figure*}

\noindent \textbf{Model Overview.}
ChartDETR follows the DETR pipeline and the details are described in Figure~\ref{fig:chart_detr}. Given a chart image, ChartDETR first applies the backbone and the transformer encoder to extract the refined visual features. Second, with the refined visual features, a novel query grouping decoder generates keypoint embeddings from a set of initialized keypoint queries and divides them into groups. Last, for each group of keypoint embeddings, the shape prediction header predicts all keypoint positions and the class probability.
% Last, the shape prediction header takes keypoint embeddings and predicts all keypoint positions and the class probability of each group. 
We train ChartDETR with the Hungarian algorithm and propose the duplicated keypoint matching to address the problem of a variable number of keypoints in line charts. \\

\noindent \textbf{Query Grouping Decoder.}
Different from most object queries, which predict a bounding box from each query, it is difficult to predict all keypoints of a shape from one query because the chart shape (\emph{e.g.}, line chart) may span a wide range. Therefore, our decoder simultaneously predicts all keypoints of $M$ shapes and divides them into groups to represent different shapes by a simple way.

Specifically, we initialize the $(M \times N)$-length keypoint queries $Q \in \mathbb{R}^{(M \times N) \times D}$, where $D$ is the dimension of the query embedding, $N$ is the keypoint number of a shape, and $M$ is the number of shapes that the model can predict. With the refined visual features from the encoder, the decoder utilizes the self/cross-attention module layer by layer and transforms the keypoint queries into the keypoint query embeddings $E \in \mathbb{R}^{(M \times N) \times D}$.

Considering that $E$ contains all keypoint embeddings of $M$ shapes and each shape has $N$ keypoints, we simply divide the sequential embeddings into $M$ groups, and the shape of $E$ changes into $M \times N \times D$. In the $m$-th group, each keypoint embedding $e_n^m$ will be fed into the keypoint regression head to predict its position. We choose the first embedding $e_0^m$ as a pivot to predict whether this keypoint group corresponds to a shape. As a result, the model simultaneously predicts $M$ shapes with $N$ keypoints for each in just one shot. \\

\noindent \textbf{Duplicated Keypoint Matching.} 
Our ChartDETR outputs $N$ keypoints for each predicted shape. For bars and pies, we can easily compare a pair of shapes between predictions and the ground truth because their shapes have the same number of keypoints, \emph{e.g.}, any one of the bars is represented with its top-left and bottom-right corners wherever it is from the predictions or the ground truth. However, it is difficult to compare a predicted line shape with a ground truth one because the number of line keypoints varies in different line charts. To address this problem, we keep all line shapes with the same number of keypoints in a simple way. 

\begin{figure}
    \centering
    \includegraphics[width=\linewidth]{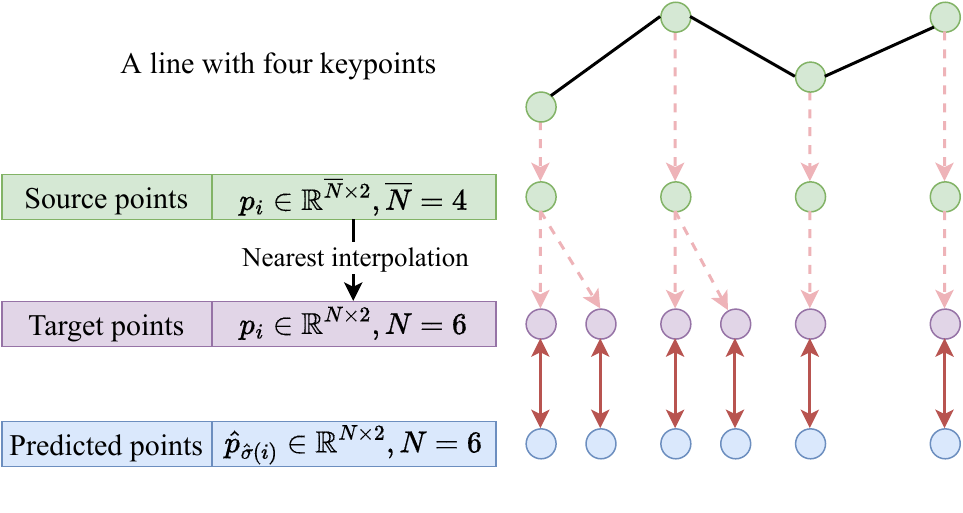}
    \caption{Duplicated keypoint matching. To compare the predicted line with the ground truth that has various numbers of keypoints, we interpolate duplicated keypoints to the ground truth with the nearest neighbors.}
    \label{fig:keypoint_matching}
    \vspace{-3mm}
\end{figure}

Specifically, we denote the ground truth keypoint set of a line as $p_i \in \mathbb{R}^{\overline{N} \times 2}$, where $\overline{N}$ is the number of ground truth keypoints, and each point has a 2D position. We interpolate duplicated points into $p_i$ by using nearest neighbors to get the updated keypoint set $p_i \in \mathbb{R}^{N \times 2}$, where $N \geq \overline{N}$ to ensure that the source keypoints will not be removed. This way, the interpolated keypoint set has the same number of keypoints with the predicted line shapes and thus can serve as the target to supervise the model training in a DETR way. We illustrate this operation in Figure~\ref{fig:keypoint_matching} for better understanding. \\

\noindent \textbf{Network Implementation.} 

\noindent - \emph{Backbone}: 
Given an input image $I \in \mathbb{R}^{H \times W \times 3} $, which can be either type of bar, line, or pie chart. We extract multi-scale feature maps from the backbone (\emph{e.g.}, ResNet-50~\cite{he2016deep}) and reduce their channels to $256$. We flatten the spatial dimensions of each feature map and concatenate them to construct the multi-scale feature representation $F$.
% with the length of $U$, $F \in \mathbb{R}^{U \times 256}$.

\noindent - \emph{Transformer encoder}: 
We follow~\cite{zhu2020deformable} to build the encoder, which has six layers, and each layer comprises a multi-scale deformable attention module and a feed-forward network. The multi-scale deformable module is designed to attend features to reference points, which is appropriate for the keypoints detection task. The scale-level embedding and the positional embedding are added to identify which feature level each feature token lies in. After that, we can obtain the refined multi-scale visual feature memory $F$.

\noindent - \emph{Transformer decoder}: 
The decoder also has six layers. Besides a multi-scale deformable self-attention module and a feed-forward network, each decoder layer also includes a multi-scale deformable cross-attention module. To pool more comprehensive content features and fully leverage the refined keypoint information, two improvements are borrowed from DINO~\cite{zhang2022dino}: (1) the mixed query selection that generates keypoint queries composed by a positional part and a content part; (2) the look forward twice approach that refines the keypoints at each layer based on the predictions from the two previous layers.

\noindent - \emph{Shape prediction header}: 
Our shape header consists of a classification head including a linear layer and a keypoint regression head composed of a multi-layer perceptron. The classification head takes the embeddings of all group pivots and outputs the confidence through the sigmoid function. The keypoint regression head takes all keypoint embeddings and predicts the offsets w.r.t. the reference points. 

\subsection{Training with Hybrid Shape Matching}
\label{subsec:matching}
DETR uses the one-to-one set matching to force a unique prediction for each ground-truth object. However, the training efficiency may be significantly reduced due to few positive samples. To leverage recent improvements of DETR variants, we apply the hybrid matching scheme~\cite{jia2022detrs} to train our ChartDETR due to its simplicity and convenience to transfer from bounding box detection to multi-shape detection. \\
% To leverage recent improvements of DETR variants, during training, we match targets with predictions in a hybrid scheme~\cite{jia2022detrs} for loss computation. Moreover, we propose the duplicated keypoint matching to address the problem of a variable number of keypoints in line charts. \\

\noindent \textbf{One-to-One Shape Matching.}
For the queries $ Q \in  \mathbb{R}^{(M \times N) \times D}$, we denote the corresponding prediction sets as $\widehat{S}=\{\hat{s}_i\}^M$. Each predicted shape $\hat{s}_i$ includes a class probability $\hat{q}_i$ and a keypoint set $\hat{p}_i$.  Let $G$ denote the target shape set. The class label and target keypoint set of each shape $g_i$ are $c_i$ and $p_i$, respectively. we perform the bipartite matching between $\widehat{S}$ and $G$ to estimate the optimal assignment $\hat{\sigma}_{one2one}$ by minimizing the matching cost:
\begin{gather}
    \hat{\sigma}_{one2one} = \argmin_{\sigma \in \mathfrak{S}_M} \sum_{i=0}^{M-1}\mathcal{L}_{match} \left (\hat{s}_{\sigma(i)}, g_i \right), \\
    \begin{split}
        \mathcal{L}_{match}(\hat{s}_{\sigma(i)}, g_i) = & -\vmathbb{1}_{\{c_i \neq \varnothing\}}\mathcal{L}_{cls} \left (\hat{q}_{\sigma(i)}, c_i \right) + \\
        & \vmathbb{1}_{\{c_i \neq \varnothing\}}\mathcal{L}_{shape} \left (\hat{p}_{\sigma(i)}, p_i \right),
    \end{split}
\end{gather}
where $\mathfrak{S}_M$ is a permutaion of $M$ shapes, $\mathcal{L}_{match}(\hat{s}_{\sigma(i)}, g_i)$ computes the classification cost $\mathcal{L}_{cls} (\cdot)$ and the shape position cost $\mathcal{L}_{shape} (\cdot)$ between the target $g_i$ and a prediction with index $\sigma(i)$. We use the focal loss\cite{lin2017focal} as $\mathcal{L}_{cls} (\cdot)$ and $\ell_1$ loss as $\mathcal{L}_{shape} (\cdot)$. With the optimal assignment $\hat{\sigma}_{one2one}$, we then compute the one-to-one matching loss with the Hungarian algorithm: 
\begin{gather}
    \mathcal{L}_{one2one} = \mathcal{L}_{Hungarian}(\widehat{S}, G ), \\
    \begin{split}
    \mathcal{L}_{Hungarian}(\widehat{S}, G ) = & \sum_{i=0}^{M-1} \left[  \mathcal{L}_{cls} \left(\hat{q}_{\hat{\sigma}_{one2one}(i)}, c_i \right) + \right. \\ 
    & \left. \vmathbb{1}_{\{c_i \neq \varnothing\}}\mathcal{L}_{shape} \left(\hat{p}_{\hat{\sigma}_{one2one}(i)}, p_i \right) \right],
    \end{split}
\end{gather}

\noindent \textbf{One-to-Many Shape Matching.}
Besides the queries $Q$ for one-to-one matching, we also maintain another group of queries $ \widetilde{Q} \in  \mathbb{R}^{(T \times N) \times D}$ for one-to-many matching, where $T$ is the number of predicted shapes for $\widetilde{Q}$. Let $\widetilde{S}=\{\tilde{s}_i\}^T$ denote the corresponding prediction sets, where each predicted shape $\tilde{s}_i$ has a class probability $\tilde{q}_i$ and a keypoint set $\tilde{p}_i$. To perform the one-to-many matching, a simple way is to repeat the ground truth $G$ for $K$ times and get an augmented target $\widetilde{G}$. We then estimate the optimal assignment and compute the one-to-many matching loss in a similar way, 
\begin{gather}
    \hat{\sigma}_{one2many} = \argmin_{\sigma \in \mathfrak{S}_T} \sum_{i=0}^{T-1}\mathcal{L}_{match} \left (\tilde{s}_{\sigma(i)}, \tilde{g}_i \right), \\
    \mathcal{L}_{one2many} = \mathcal{L}_{Hungarian}(\widetilde{S}, \widetilde{G} ).
\end{gather}

We parallelly process the one-to-one and one-to-many branches by a masked multi-head self-attention. The total loss is the combination of the above two losses,
\begin{equation}
    \mathcal{L}_{total} = \mathcal{L}_{one2one} + \mathcal{L}_{one2many}.
\end{equation}

\begin{table*}[t]
    \renewcommand\arraystretch{1.1}
    % \small
    \centering
    \caption{Pie and bar detection performance on  ExcelChart400k.}
    \begin{tabular}{l|l|l|c|c}
    \hline
     Method & Backbone & Framework & Pie Score & Bar Score\\
    \hline
     \multicolumn{5}{c}{Human Pose Estimation Methods} \\
    \hline
     HRNet$^{\dagger}$ & HRNet-w32 & Top-down Keypoint Detection & 0.952 & 0.910 \\
     HigherHRNet & HRNet-w32 & Bottom-up Keypoint Detection & 0.938 &  0.615 \\
     PETR & ResNet-50 & One-stage Keypoint Detection & 0.956 & 0.902 \\
    \hline
    \multicolumn{5}{c}{Chart Recognition Methods} \\
    \hline
     Revision & -- & Rule-based & 0.838 & 0.582 \\
     RotationRNN & ResNet-50 & Object Detection & 0.797 & - \\
     Faster-RCNN & ResNet-50 & Object Detection & - & 0.802 \\
     ChartOCR & Hourglass & Bottom-up Keypoint Detection & 0.918 & 0.919\\
    \hline
     ChartDETR & ResNet-50 & One-stage Keypoint Detection & 0.963 & 0.904 \\
     ChartDETR & HRNet-w32 & One-stage Keypoint Detection & \textbf{0.975} & \textbf{0.923} \\ 
    %  ChartDETR & 0.881 & 0.904\\
    % \hline
    %  ChartDETR (w/ regrouping) & \textbf{0.963} & - \\
     \hline
     \multicolumn{5}{l}{${\dagger}$: The results of HRNet are based on the ground truth bounding boxes of shape keypoints.} 
    \end{tabular}
    \label{tab:pie_bar_det}
    % \vspace{-3mm}
\end{table*}

\begin{table*}[t]
    \renewcommand\arraystretch{1.1}
    % \small
    \centering
    \caption{Line detection performance on Adobe Synthetic (AS), LINEEX430k (LE), and ExcelChart400k (EC).}
    \begin{tabular}{l|l|l|l|c|c|c|c|c|c|c|c|c}
    \hline
    % \multicolumn{7}{c}{Adobe Synthetic} \\
    % \hline
    % \multirow{2}{*}{Method} & \multicolumn{3}{c|}{$OKS_{str}$} & \multicolumn{3}{c}{$OKS_{rel}$} \\
    % \cline{2-7}
    % & R & P & F & R & P & F \\
    %  \multirow{2}{*}{Metric} & \multirow{2}{*}{Method} & \multicolumn{3}{c|}{AS~\cite{davila2019icdar}} & \multicolumn{3}{c|}{LE~\cite{shivasankaran2023lineex}} & \multicolumn{3}{c}{EC~\cite{luo2021chartocr}} \\
    \multirow{2}{*}{Metric} & \multirow{2}{*}{Method} & \multirow{2}{*}{Backbone} & \multirow{2}{*}{Framework} & \multicolumn{3}{c|}{AS} & \multicolumn{3}{c|}{LE} & \multicolumn{3}{c}{EC} \\
    \cline{5-13}
     & & & & R & P & F1 & R & P & F1 & R & P & F1\\
    \hline
    %  \multirow{3}{*}{$OKS_{str}$} & ChartOCR~\cite{luo2021chartocr} & Hourglass & Bottom-up KD & 0.76 & 0.72 & 0.71 & 0.71 & 0.90 & 0.78 & \textbf{0.85} & \textbf{0.98} & \textbf{0.90} \\
    \multirow{3}{*}{$OKS_{str}$} & ChartOCR & Hourglass & Bottom-up & 0.76 & 0.72 & 0.71 & 0.71 & 0.90 & 0.78 & \textbf{0.85} & \textbf{0.98} & \textbf{0.90} \\
                                % & LINEEX~\cite{shivasankaran2023lineex} &  & & 0.91 & 0.54 & 0.64 & 0.86 & 0.84 & 0.83 & 0.84 & 0.80 & 0.78 \\
                                & LINEEX & Transformer & Bottom-up & 0.91 & 0.54 & 0.64 & 0.86 & 0.84 & 0.83 & 0.84 & 0.80 & 0.78 \\
                                & ChartDETR & ResNet-50 & One-stage & \textbf{0.97} & \textbf{0.99} & \textbf{0.98} & \textbf{0.92} & \textbf{0.98} & \textbf{0.95} & 0.84 & 0.92 & 0.86 \\
     \hline
     \multirow{3}{*}{$OKS_{rel}$} & ChartOCR & Hourglass & Bottom-up & 0.78 & 0.80 & 0.76 & 0.74 & 0.97 & 0.83 & 0.85 & \textbf{0.98} & \textbf{0.90} \\
                                & LINEEX & Transformer & Bottom-up & 0.93 & 0.76 & 0.81 & 0.85 & 0.92 & 0.87 & 0.85 & 0.90 & 0.85\\
                                & ChartDETR & ResNet-50 & One-stage & \textbf{0.97} & \textbf{0.99} & \textbf{0.98} & \textbf{0.93} & \textbf{0.98} & \textbf{0.95} & \textbf{0.85} & 0.95 & 0.88 \\
     \hline
    \end{tabular}
    \label{tab:line_det}
    % \vspace{-3mm}
\end{table*}

\section{Experiment}
\subsection{Setup}
\noindent \textbf{Datasets.} 
We employ our method on three types of charts (\emph{i.e.}, bar, line, and pie) and evaluate on Adobe Synthetic~\cite{davila2019icdar}, LINEEX430k~\cite{shivasankaran2023lineex}, and ExcelChart400k~\cite{luo2021chartocr}. \\
% Their statistics are listed in Fig.~\ref{fig:stat_dataset}. 

\noindent \textbf{Metrics.} 
For bar and pie charts, we follow~\cite{luo2021chartocr} to compute the detection score. For a line chart, we follow ~\cite{shivasankaran2023lineex} to evaluate the line detection results based on the strict and relaxed object keypoints similarities, which are denoted as $OKS_{str}$ and $OKS_{rel}$, respectively. \\

\noindent \textbf{Baseline methods.}
We compare our ChartDETR with a wide range of methods, including Revision~\cite{savva2011revision}, RotationRNN and Faster-RCNN~\cite{liu2019data}, HRNet~\cite{sun2019deep}, HigherHRNet~\cite{cheng2020bottom}, ChartOCR~\cite{luo2021chartocr}, PETR~\cite{shi2022end}, and LINEEX~\cite{shivasankaran2023lineex}. \\

\noindent \textbf{Implementation.} 
For all experiments, we use ResNet-50~\cite{he2016deep} as the backbone if not specially specified. Considering the limited GPU memory, we always keep $T = 3 \times M$ for one-to-many matching.
% and choose the training line charts of less than 64 keypoints per line.

\subsection{Pie Detection}
We evaluate our ChartDETR for pie detection on ExcelChart400k and present the experimental results in Table~\ref{tab:pie_bar_det}. Compared with chart recognition methods, our ChartDETR can directly output a set of pie sectors with three keypoints without extra post-processing. When adopting HRNet as the backbone, ChartDETR achieves a 0.975 score that improves the recent best result (ChartOCR) by 6.2\%, obtaining a new state-of-the-art result. Compared with recent human pose estimation methods, ChartDETR-ResNet-50 achieves a 0.963 score higher than PETR, which is also a DETR-like method. We visualize some challenging samples in Figure~\ref{fig:all_vis}. We see that ChartDETR can effectively predict the invisible center point of the special donut type and two close endpoints of a pie sector that is only 1\% of the whole circle. 
%These results demonstrate the effectiveness of the proposed ChartDETR for pie shape detection. 
%Moreover, ChartDETR can directly output a set of pie sectors, which is simpler than previous methods.

\subsection{Bar Detection}
\label{subsec:bar_det}
We present bar detection results on ExcelChart400k in Table~\ref{tab:pie_bar_det}. Our ChartDETR-HRNet achieves a 0.923 score, outperforming all other baseline methods. HigherHRNet has a low score because it struggles to distinguish overlapped keypoints between neighbor bars. Both ChartDETR and PETR are the DETR-like framework. When adopting ResNet-50 as the backbone, they have comparable results (0.904: 0.902) and fall behind ChartOCR. The main reason is that ChartOCR utilizes a high-resolution backbone (\emph{e.g.}, Hourglass Net~\cite{newell2016stacked}), while ChartDETR uses a low-resolution backbone, facing the challenge to predict small bars, as shown in Figure~\ref{fig:all_vis}. When replacing ResNet-50 with a high-resolution backbone, \emph{e.g.} HRNet, the performance can be further improved (0.904 $\rightarrow$ 0.923). 

\subsection{Line Detection}
\begin{figure*}[t]
    \centering
    \includegraphics[width=\linewidth]{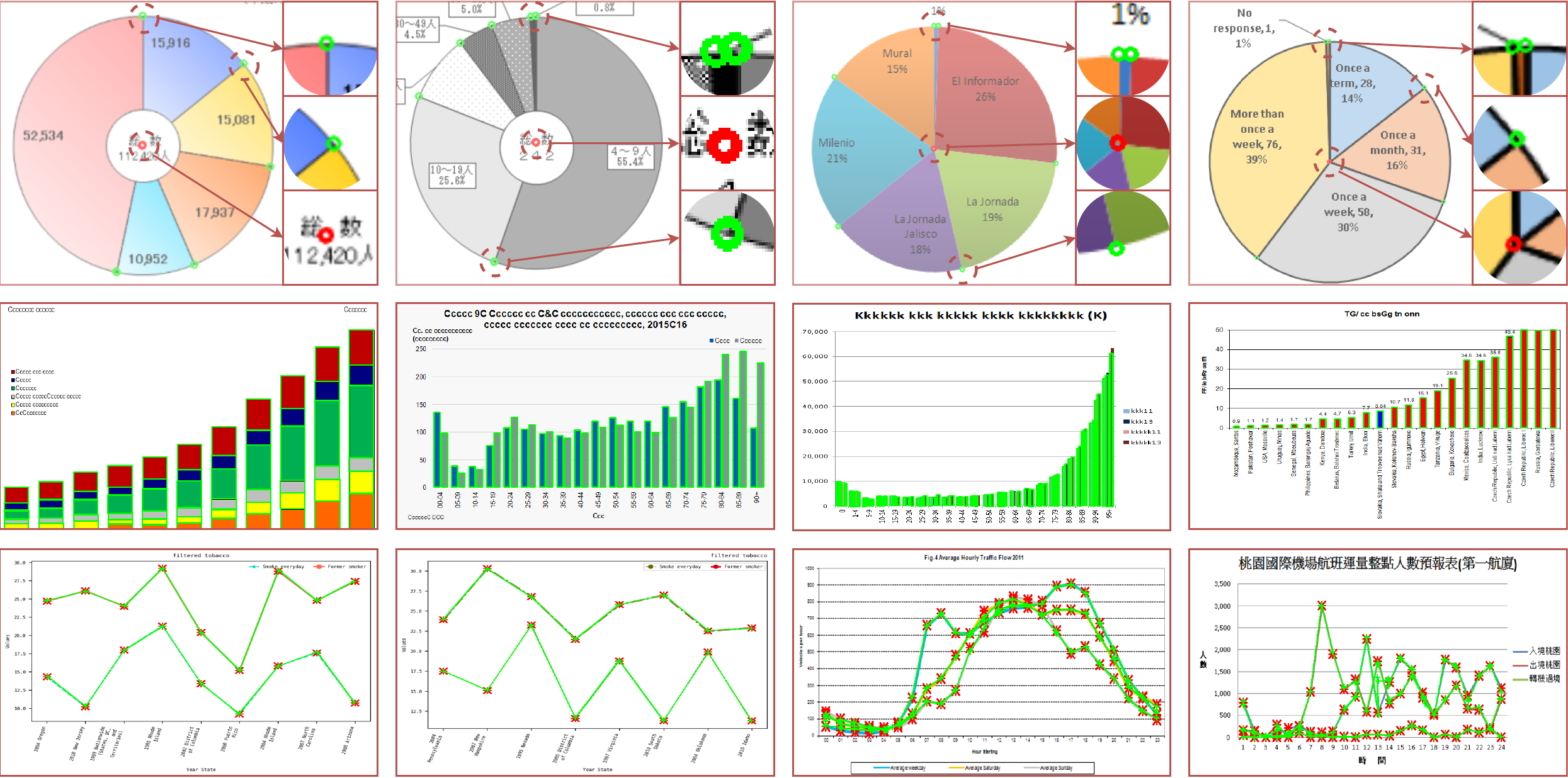}
    \caption{Visualization results of ChartDETR. First row: pie charts where green circles (\textcolor{green}{$\circ$}) are predicted endpoints and red circles (\textcolor{red}{$\circ$}) are predicted center points. Second row: bar charts where green boxes (\textcolor{green}{$\square$}) are predicted bars. Third row: line charts where predicted keypoints (\textcolor{green}{$+$}) are connected with green lines and red asterisks (\textcolor{red}{$\ast$}) are the target keypoints.}
    \label{fig:all_vis}
    \vspace{-3mm}
\end{figure*}

Line detection is the most challenging among the three chart types due to the irregular and various keypoints distribution. We evaluate our methods for line detection on three different datasets and present the results in Table~\ref{tab:line_det}. On Adobe Synthetic and LINEEX430k, our ChartDETR achieves 0.98 F1 and 0.95 F1 for the $OKS_{str}$ metric, improving recent best results by 53.1\% and 14.4\%, respectively. Compared with previous methods, the proposed method has a balanced performance for both $OKS_{str}$ and $OKS_{rel}$ metrics. That demonstrates the predicted keypoints closely surround their target points. On ExcelChart400k, though our ChartDETR achieves 0.86 F1 and 0.88 F1 for $OKS_{str}$ and $OKS_{rel}$ metrics, lowering than the previous best results, it has a comparable recall. It is noted that our method predicts superfluous keypoints. When a line has few knee points and approximates to a straight line, instead of clustering around the knee points, the predicted superfluous points tend to arrange along the line and become false positive samples, which results in low precision. However, compared with these methods, our ChartDETR can directly output a set of lines without extra post-processing. We present visualization results in Figure~\ref{fig:all_vis}.
% However, these previous methods have complex hand-craft algorithms to filter and group keypoints. Compared with these methods, our ChartDETR can directly output a set of lines without extra post-processing. We present visualization results in Fig.~\ref{fig:all_vis}.

\subsection{Ablation Study}
We present ablation experiments to study the influence of several key parameters for line charts in Adobe Synthetic. Different backbones are also investigated for bar detection on  ExcelChart400k.
% For all experiments, models are trained for $25$ epochs. 
We set $K=3$ by default for the one-to-many matching if not specially specified. \\

\begin{table}
    \small
    \centering
    \caption{Influence of $N$ and $M$ for line detection on Adobe Synthetic. $\times x$ indicates the $x$ times of the default value.}
    \begin{tabular}{c|c|c|c|c}
    \hline
     $N$ & Default (14) & $\times 2$ & $\times 3$ & $\times 4$ \\
    \hline
    %  $OKS_{str}$ F1 & 0.9755 & 0.9422 & 0.9072 & 0.9101 \\
     $OKS_{str}$ F1 & 0.975 & 0.942 & 0.907 & 0.910 \\
    \hline
     $M$ & Default (4) & $\times 2$ & $\times 3$ & $\times 4$ \\
    \hline
    %  $OKS_{str}$ F1 & 0.9742 & 0.9807 & 0.9596 & 0.9437 \\
     $OKS_{str}$ F1 & 0.974 & 0.980 & 0.959 & 0.943 \\
    \hline
    \end{tabular}
    \label{tab:ablation_src_inter}
    % \vspace{-3mm}
\end{table}

\noindent \textbf{The Influence of $N$.}
To compare the predicted line keypoints with the ground truth that has a variable number of keypoints, we interpolate duplicated keypoints into the ground truth by using the nearest neighbors to get the training target. The default value of $N$ is the maximum number of line keypoints for a chart dataset, \emph{i.e.}, 14 for Adobe Synthetic. We conduct experiments to study the influence of $N$ in Table~\ref{tab:ablation_src_inter}. The model achieves the best result when setting $N$ with the default value. Interpolating more keypoints has no help to improve the model performance. \\

\noindent \textbf{The Influence of $M$.}
$M$ is an important hyper-parameter that determines the total number of keypoint queries and how many shapes the model can predict. Recent DETR variants~\cite{zhang2022dino,jia2022detrs} achieve better results with more queries. We conduct experiments to select a suitable value for $M$ in Table~\ref{tab:ablation_src_inter}. We set $M=4$ as the default value, which is the maximum number of shapes per image on Adobe Synthetic. Increasing $M$ does not always result in a better performance. The model achieves the best result when $M = 8$. \\

% \begin{figure}
%     \centering
%     \includegraphics[width=\linewidth]{figs/pie_vis_v2.pdf}
%     \caption{Visualization results of ChartDETR for pie detection on ExcelChart400k test set. First row: donuts, a special pie chart type. Second row: hard samples with two close endpoints. Green circles (\textcolor{green}{$\circ$}) are predicted endpoints on the arc and red circles (\textcolor{red}{$\circ$}) are predicted center points.}
%     \label{fig:pie_vis}
% \end{figure}

% \begin{figure}
%     \centering
%     \includegraphics[width=\linewidth]{figs/bar_vis.pdf}
%     \caption{Visualization results of ChartDETR for bar detection on ExcelChart400k test set. First row: easy samples with clean layouts. Second row: hard samples with small bars and small aspect ratios. Green boxes (\textcolor{green}{$\square$}) are predicted bars.}
%     \label{fig:bar_vis}
% \end{figure}

% \begin{figure}
%     \centering
%     \includegraphics[width=\linewidth]{figs/line_vis_v2.pdf}
%     \caption{Visualization results of ChartDETR for line detection. First row: easy samples with few keypoints and separate lines on Adobe Synthetic test set. Second row: hard samples with many keypoints and cross lines. Green lines are line segments connected by predicted keypoints (green crosses \textcolor{green}{$+$}). Red asterisks (\textcolor{red}{$\ast$}) are the ground truth keypoints.}
%     \label{fig:line_vis}
% \end{figure}

\noindent \textbf{The Influence of $K$.}
During training, we repeat the ground truth $K$ times to perform one-to-many matching followed by H-DETR. We study the influence of $K$ with different values in Table~\ref{tab:ablation_k}. We set $N=14$ and $M=8$ on Adobe Synthetic according to previous experiments. When $K=0$, only the one-to-one matching branch participates in the training, which results in a 0.951 score for $OKS_{str}$ F1 metric. We observe that the model significantly benefits from the one-to-many matching scheme when $K \geq 3$. The model achieves the best result when $K = 3$, which is 0.029 higher than the performance of the one-to-one matching only branch. The experimental results demonstrate the hybrid matching scheme also works for line detection. \\

% \begin{table}[t]
%     \centering
%     \caption{The choice of $M$ for line detection on Adobe Synthetic. $\times x$ indicates the $x$ times of the default value.}
%     \begin{tabular}{c|c|c|c|c}
%     \hline
%      $M$ & Default (4) & $\times 2$ & $\times 3$ & $\times 4$ \\
%     \hline
%     %  $OKS_{str}$ F1 & 0.9742 & 0.9807 & 0.9596 & 0.9437 \\
%      $OKS_{str}$ F1 & 0.974 & 0.980 & 0.959 & 0.943 \\
%     \hline
%     \end{tabular}
%     \label{tab:ablation_m}
% \end{table}

\begin{table}
    \small
    \centering
    \caption{The influence of $K$ for on-to-many matching on Adobe Synthetic. }
    \begin{tabular}{c|c|c|c|c|c|c}
    \hline
     $K$ & 0 & 1 & 2 & 3 & 4 & 5 \\
    \hline
    %  $OKS_{str}$ F1 & 0.9515 & 0.9562 & 0.9468 & 0.9807 & 0.9643 & 0.9696 & 0.9436 \\
     $OKS_{str}$ & 0.951 & 0.956 & 0.946 & 0.980 & 0.964 & 0.969 \\
    %  & 0.980 & 0.964 & 0.969  \\
    % \hline
    % $K$ & 3 & 4 & 5 \\
    % \hline
    % $OKS_{str}$ & 0.980 & 0.964 & 0.969 \\
    \hline
    \end{tabular}
    \label{tab:ablation_k}
\end{table}

\begin{table}
    \small
    \centering
    \caption{Bar detection with different backbones.}
    \begin{tabular}{l|c|l|c}
    \hline
     Backbone & Bar Score & Backbone & Bar Score\\
    \hline
     ResNet-50 & 0.904 & ResNet-101 & 0.905\\
    \hline
     Swin-tiny & 0.917 & Swin-small & 0.918\\
    \hline
     HRNet-w18 & 0.913 & HRNet-w32 & 0.923 \\
    \hline
    \end{tabular}
    \label{tab:ablation_backbone}
    \vspace{-3mm}
\end{table}

\noindent \textbf{Different Backbones.}
In Section~\ref{subsec:bar_det}, we show that the model can be improved by replacing ResNet-50 with HRNet. To further study the effect of backbones on ChartDETR, we conduct experiments with different backbones for bar detection in Table~\ref{tab:ablation_backbone}. The results of ResNet and Swin Transformer~\cite{liu2021swin} demonstrate that the depth and scale of backbones have a limited effect on improving performance. The high-resolution network HRNet achieves significant improvements, which demonstrates the main challenge of bar detection is the small and crowded bars. 

% \begin{figure*}[t]
%     \centering
%     \includegraphics[width=\linewidth]{figs/all_vis.pdf}
%     \caption{Visualization results of ChartDETR. First row: pie charts where green circles (\textcolor{green}{$\circ$}) are predicted endpoints and red circles (\textcolor{red}{$\circ$}) are predicted center points. Second row: bar charts where green boxes (\textcolor{green}{$\square$}) are predicted bars. Third row: line charts where predicted keypoints (\textcolor{green}{$+$}) are connected with green lines and red asterisks (\textcolor{red}{$\ast$}) are the target keypoints.}
%     \label{fig:all_vis}
%     \vspace{-3mm}
% \end{figure*}

\section{Conclusion}
This paper introduces a novel transformer-based multi-shape detector for chart recognition, \emph{i.e.}, ChartDETR. The proposed method transforms multi-shape detection into a set prediction issue by incorporating query groups and duplicated keypoint matching into a unified framework. This approach enables ChartDETR to effectively detect data elements of various shapes in charts without requiring a change in network architecture. Compared to other methods, ChartDETR is straightforward and eliminates the need for post-processing. Experimental results show the effectiveness of ChartDETR on three types of charts.

\bibliography{aaai24}

\end{document}